\documentclass[10pt, a4paper]{article}

\usepackage{lrec-coling2024} 
\usepackage{makecell}
\usepackage{multirow}
\usepackage{rotating}
\usepackage{graphicx}
\usepackage{subfigure}
\usepackage{booktabs}

\title{Exploring Boundaries and Intensities in Offensive and Hate Speech: Unveiling the Complex Spectrum of Social Media Discourse}

\name{\normalsize Abinew Ali Ayele$^{1,2}$, Esubalew Alemneh Jalew$^{2}$, Adem Chanie Ali$^{2}$,  \\
\textbf{\normalsize Seid Muhie Yimam$^{1}$, Chris Biemann$^{1}$}\\\\
\footnotesize
\address {$^1$ Universität Hamburg, Germany, $^2$ Bahir Dar University, Ethiopia
}}

\abstract{
The prevalence of digital media and evolving sociopolitical dynamics have significantly amplified the dissemination of hateful content. Existing studies mainly focus on classifying texts into binary categories, often overlooking the continuous spectrum of offensiveness and hatefulness inherent in the text.
In this research, we present an extensive benchmark dataset for Amharic, comprising 8,258 tweets annotated for three distinct tasks: \emph{category classification}, \emph{identification of hate targets}, and \emph{rating offensiveness and hatefulness intensities}. 
Our study highlights that a considerable majority of tweets belong to the \emph{less offensive} and \emph{less hate} intensity levels, underscoring the need for early interventions by stakeholders. The prevalence of \emph{ethnic} and \emph{political} hatred targets, with significant overlaps in our dataset, emphasizes the complex relationships within Ethiopia's sociopolitical landscape.
We build classification and regression models and investigate the efficacy of models in handling these tasks. Our results reveal that hate and offensive speech can not be addressed by a simplistic binary classification, instead manifesting as variables across a continuous range of values. The Afro-XLMR-large model exhibits the best performances achieving F1-scores of 75.30\%, 70.59\%, and 29.42\% for the category, target, and regression tasks, respectively. The 80.22\% correlation coefficient of the Afro-XLMR-large model indicates strong alignments. 
 \\ \newline  }

\begin{document}

\maketitleabstract

\section{Introduction}

In the world of rapid innovations, the prevalence and influence of social media persistently expand, along with the diverse array of online content crafted by a multitude of contributors, which has become readily available for consumption and engagement \citep{sazzed-2023-discourse}. Remarkably, over 60\% of the world's population is actively participating in social media. However, social media platforms have become the main places for the dissemination and proliferation of hate speech \citep{unesco-Bran2023,mathew2021hatexplain,davidson2017automated,waseem2016hateful,ayele-etal-2023-exploring}. The ease of communication and the global reach of these platforms have enabled users to spread hateful and offensive content aggressively in wider circles \citep{zufall-etal-2022-legal}. The anonymity of online users on social media granted hateful message propagators to spread toxic content by hiding themselves behind their digital screens \citep{unesco-Bran2023,DBLP:journals/jair/KiritchenkoNF21,zufall-etal-2022-legal}. Hate speech on social media can take various forms, including discriminatory language, threats, harassment, and the incitement of violence against specific individuals or groups of communities \citep{mathew2021hatexplain,davidson2017automated,ayele-etal-2023-multilingual}. This online hate speech can have real-world consequences, contributing to social divisions, fueling hostility, and inciting violence in some circumstances \citep{abraha2017examining,muhie2019analysis}. As a result, social media companies, policymakers, and researchers are increasingly focused on developing strategies to detect, combat, and mitigate the impact of hate speech on these platforms without compromising the principles of freedom of speech and user safety \citep{pavlopoulos-etal-2017-deeper,ayele-etal-2023-multilingual}.

For the past couple of years, there has been increasing attention and interest in exploring hate speech among researchers from diverse academic disciplines, including social science, psychology, media and communications studies, and computer science \cite{tontodimamma2021thirty,davidson2017automated,mathew2021hatexplain,davidson-etal-2019-racial,chekol2023social,ayele-etal-2023-exploring}. 

Many studies, including those by \citet{davidson2017automated,fortuna-etal-2020-toxic,waseem2016hateful,mathew2021hatexplain,plaza-del-arco-etal-2023-respectful,clarke-etal-2023-rule,caselli-van-der-veen-2023-benchmarking} and others, adopt a binary approach to hate speech classification. These works aim to distinguish and label content as either hate or non-hate. Nevertheless, this binary viewpoint lacks the capacity to capture the diverse and context-dependent features of hate speech, which resist easy classification. We posit that hate speech classification demonstrates a spectrum of continuity \cite{Bahador2023}. In contemporary studies, there has been a recognition of this limitation by prompting a shift towards adopting multifaceted methodologies to gain a better understanding of the nature, dimension, and intensity of hate speech \cite{beyhan-etal-2022-turkish,sachdeva-etal-2022-measuring}. This further enhances hate speech detection capabilities and employs more effective mitigation strategies to tackle its propagation on social media and its impact on the physical world. 

Studies on hate speech in low-resource languages, particularly Amharic, such as those conducted by \citet{abebaw2022design,mossie2018social,ayele20225js, Tesfaye2020,ayele-etal-2023-exploring}, predominantly concentrated on the detection of hate speech as a binary concept, overlooking its varying levels of intensities.  

In this study, our focus extends beyond the binary approach to include the varied intensities of hate and offensive speech. For the intensity rating approach, we adopt the Likert rating scale during annotation. Likert rating scale is a commonly used tool to measure attitudes, opinions, or perceptions of respondents towards a particular subject, where respondents are asked to choose the options that best reflects their viewpoint for each item  \cite{subedi2016using}. Likert rating scale provides a quantitative measurement of qualitative data, which helps researchers to analyze attitudes or opinions in a structured and comparable manner \cite{joshi2015likert}. 

The dataset was collected from X, formerly Twitter and annotated a total of 8.3k tweets. Five native Amharic speakers individually provided annotations for each tweet. Our annotations covered three distinct types: \textbf{category}, \textbf{target}, and \textbf{intensity level}.

 In the \textbf{category} type of annotations, we requested annotators to classify each tweet into specific categories. These categories include:
\begin{enumerate}
    \item \textbf{Hate}: Tweets that promote prejudice, discrimination, hostility, or violence against individuals or groups targeting their group identities to marginalize or harm them.
    \item \textbf{Offensive}: Tweets that are likely to cause discomfort, annoyance, or distress to people, but do not target any of their group identities.
   \item \textbf{Normal}: Tweets that do not contain any hate or offensive language and are considered within the boundaries of acceptable and respectful discourse.
    \item \textbf{Indeterminate}: This consists of tweets that are challenging to categorize due to various reasons, such as tweets that contain mixed languages, and typographical errors. It also includes tweets that are unclear or incomprehensible to determine its content accurately.
\end{enumerate}
The \textbf{target} annotation type involves identifying the specific groups, individuals, or communities who are the recipients of the hate speech within the tweet. This process aids in understanding the intended targets of the harmful content, providing insights into the context and potential impact.

Lastly, the \textbf{intensity level} annotation type is a valuable measure for assessing the intensities of hate and offensive speech. It provides a means to measure where a tweet falls along the spectrum of harm, from milder instances to more severe cases. This type of annotation aids in understanding the varying degrees of harm and evaluating the subtle nature of such content.

The following are the main research questions that we address in this paper: 
\begin{itemize}
    \item \textbf{RQ-1: } Do hate and offensive speech represent discrete binary categories, or exist on a continuous spectrum of varying intensities?
    \item \textbf{RQ-2: } What is the extent to which hate speech specifically targets certain groups of the population? and,  
    \item \textbf{RQ-3:} What is the occurrence and nature of tweets containing hate speech directed towards multiple target groups?
\end{itemize}

The main contributions of this study include the following but not limited to: 
\begin{enumerate}
    \item Presenting a benchmark dataset for hate speech category and target detection tasks, supplemented with intensity level ratings, 
    \item Providing comprehensive annotation guidelines for hate speech categories, targets, and approaches to measure the intensity of offensiveness and hatefulness, and 
    \item Developing classification and regression models for predicting hate intensity levels and detecting hate speech and its targets. 
\end{enumerate}

Despite focusing on Amharic, the outlined approach can be further extended to other languages and cultural contexts. 


\section{Related Works}
There is no clear and simple demarcation between hate speech, offensive speech, and protected free speech due to its complex nature. The complexity arises from the subjective nature of the offense, contextual variability, diversity of intent, varying degrees of harm, and variations in legal definitions \cite{madukwe-etal-2020-data,ayelechallenges}. Recognizing this complexity is important for balancing the protection of free speech rights with the need to address and mitigate harmful content effectively. This necessitates a holistic approach to be employed in determining the nature and consequences of such speech by considering the intent, impact, cultural context, and legal frameworks \cite{zufall-etal-2022-legal,beyhan-etal-2022-turkish,chandra-etal-2020-abuseanalyzer}.

Over the past several years, a lot of research attempts have been dedicated to exploring and analyzing hate speech using social media data. However, the majority of these studies approached hate speech detection and classification tasks as a binary categorization or dissecting it into three or four distinct classes. For instance, \citet{davidson2017automated, mathew2021hatexplain,ousidhoum-etal-2019-multilingual,waseem2016hateful,sigurbergsson-derczynski-2020-offensive,clarke-etal-2023-rule} are among the studies conducted for resourceful languages that focused on detecting hate speech and its targets. \citet{clarke-etal-2023-rule} and \citet{mathew2021hatexplain} attempted a bit deeper study and investigated explainable hate speech detection approaches beyond detecting its presence in a text. \citet{kennedy-etal-2020-contextualizing} studied hate speech by contextualizing classifiers with explanations that encourage models to learn from the context. \citet{ocampo-etal-2023-depth} explored the detection of implicit expressions of hatred, highlighting the complexity of the task and underscoring that hate speech is not yet well studied.

Hate speech detection studies conducted so far in the Amharic language also approach the problem as a binary classification task. For instance, \citet{mossie2018social, Defersha2021, abebaw2022design, Tesfaye2020} investigated Amharic hate speech as a binary hate and non-hate class, and \citet{mossie2020vulnerable} identified similar binary label categories, but further explored targeted communities. \citet{ayele20225js} explored Amharic hate speech in four categories such as hate, offensive, normal, and unsure, and \citet{ayele-etal-2023-exploring} employed similar categories except the exclusion of the unsure class in the latter study. 
In addition to textual studies, a few multimodal research attempts for Amharic such as \citet{degu2023amharic,debele2022multimodal} explored Amharic hate speech using meme text extracts and audio features, treating the task as a discrete binary task.

Recent studies indicated that hate and offensive speeches are not simple binary concepts, rather they exist on a continuum, with varying degrees of intensity, harm, and offensiveness \citep{Bahador2023,sachdeva-etal-2022-measuring}. In practical scenarios, hate speech exhibits a wide spectrum, encompassing mild stereotyping on one end and explicit calls for violence against a specific group on the other \citep{beyhan-etal-2022-turkish}.
 \citet{demus-etal-2022-comprehensive} explored hate speech categories, targets, and sentiments in two or three discrete categories while analyzing the toxicity of the message using the Likert scale ratings of 1-5 to show the potential of a message to ”poison” a conversation. 

 The study by \citet{chandra-etal-2020-abuseanalyzer} investigated the intensity of online abuse by classifying it into three separate discrete labels, namely 1) biased attitude, 2) act of bias and discrimination, and 3) violence and genocide. The annotators chose among these labels and employed the majority voting scheme for the gold labels. This online abuse intensity study employed the classical categorical approach which is a binary perspective and failed to represent the diverse fine-grained contexts in a spectrum of continuum values.  

In this study, we aim to explore the extent of offensiveness and hatefulness intensities of tweets on a rating scale of 1-5, and 0 representing normal tweets.

\section{Data Collection and Annotation}

This section presented the descriptions of data collection and annotation procedures.  

\subsection{Data Collection}

The dataset has been collected from Twitter/X spanning over 15 months since January 1, 2022. During this time, a multitude of highly controversial dynamics were occurring within the complex sociopolitical landscape of Ethiopia. Over 3.9M tweets that are written in Amharic Fidäl script were crawled, and further filtered by removing retweets, and the tweets that are written in languages other than Amharic. We used different data selection strategies such as hate and offensive lexicon entries, and the inclusion of seasons in which controversial social and political events happened. 

\subsection{Data Annotation}


\subsubsection{Overall Annotation Procedures}

\begin{figure}[!ht]
\begin{center}
\includegraphics[scale=0.29]{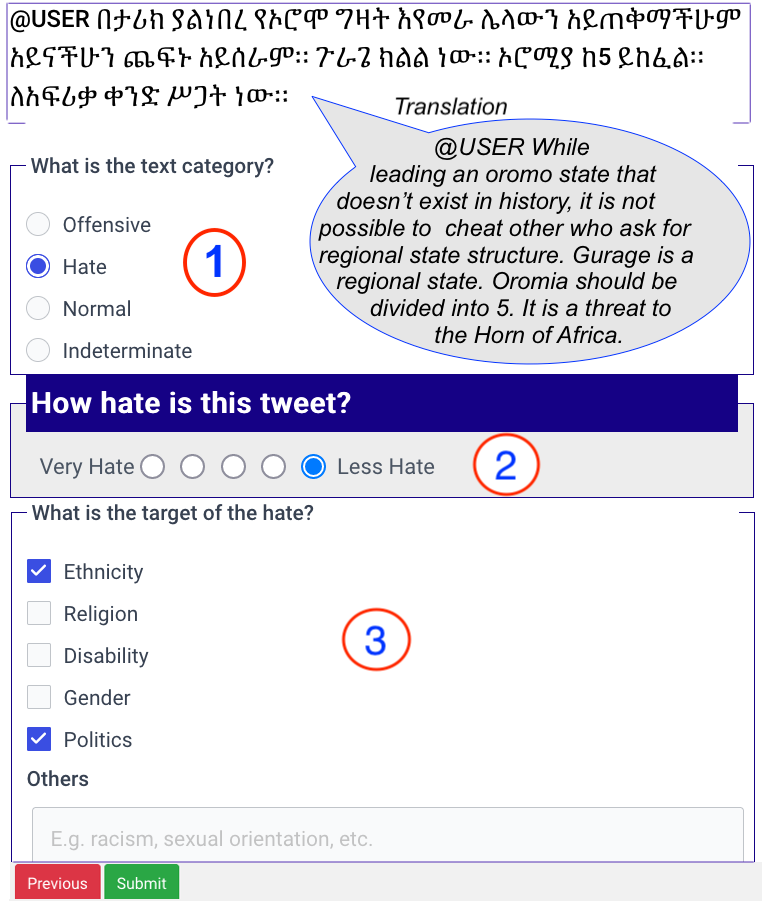} 
\caption{Potato GUI for the three types (1 - category, 2 - intensity, and 3 - target) of annotation tasks.}
\label{fig-anno-interface}
\end{center}
\end{figure}

We customized and employed the Potato-POrtable Text Annotation TOol\footnote{\url{https://github.com/davidjurgens/potato}} for the data annotation. Annotators were provided annotation guidelines, took hands-on practical training, completed independent sample test tasks, and participated in group evaluation of independent sample tests they completed. A total of 8.3k tweets are annotated into \textbf{hate}, \textbf{offensive}, \textbf{normal}, and \textbf{indeterminate} classes as shown in Table \ref{label-distribution}. Besides, annotators were requested to identify the targets of hateful tweets and also indicate their ratings of the extent of hatefulness and offensiveness intensities of tweets on a 5-point Likert scale as indicated in Figure \ref{fig-anno-interface}. The entire annotation process consists of a pilot round and five subsequent batches for the primary task annotations. 
Each tweet is annotated by 5 independent annotators, and the gold labels are determined with a majority voting scheme. A Fleiss' kappa score of 0.49 is achieved among the five annotators. We compensated annotators with a payment of \$0.03 per tweet, roughly 180 ETB per hour on average, nearly the same as the hourly wage of a Master's degree holder in Ethiopia. 

\subsubsection{Backgrounds of Annotators}
A total of 11 Amharic native speakers, 5 female and 6 male annotators, were engaged in the annotation task, representing a diverse range of ethnic, religious, gender, and social backgrounds. Annotators comprised of 6 MSc graduates and 5 MSc students from both Natural and Social Science disciplines. 

Table \ref{tab-dataset_example} presented examples, which showed the structure of the annotated dataset for the three types of annotations; namely category, hatred target and intensity (hatefulness and offensiveness) annotations. 
\begin{table*}
\begin{center}
\includegraphics[width=1.0\linewidth]{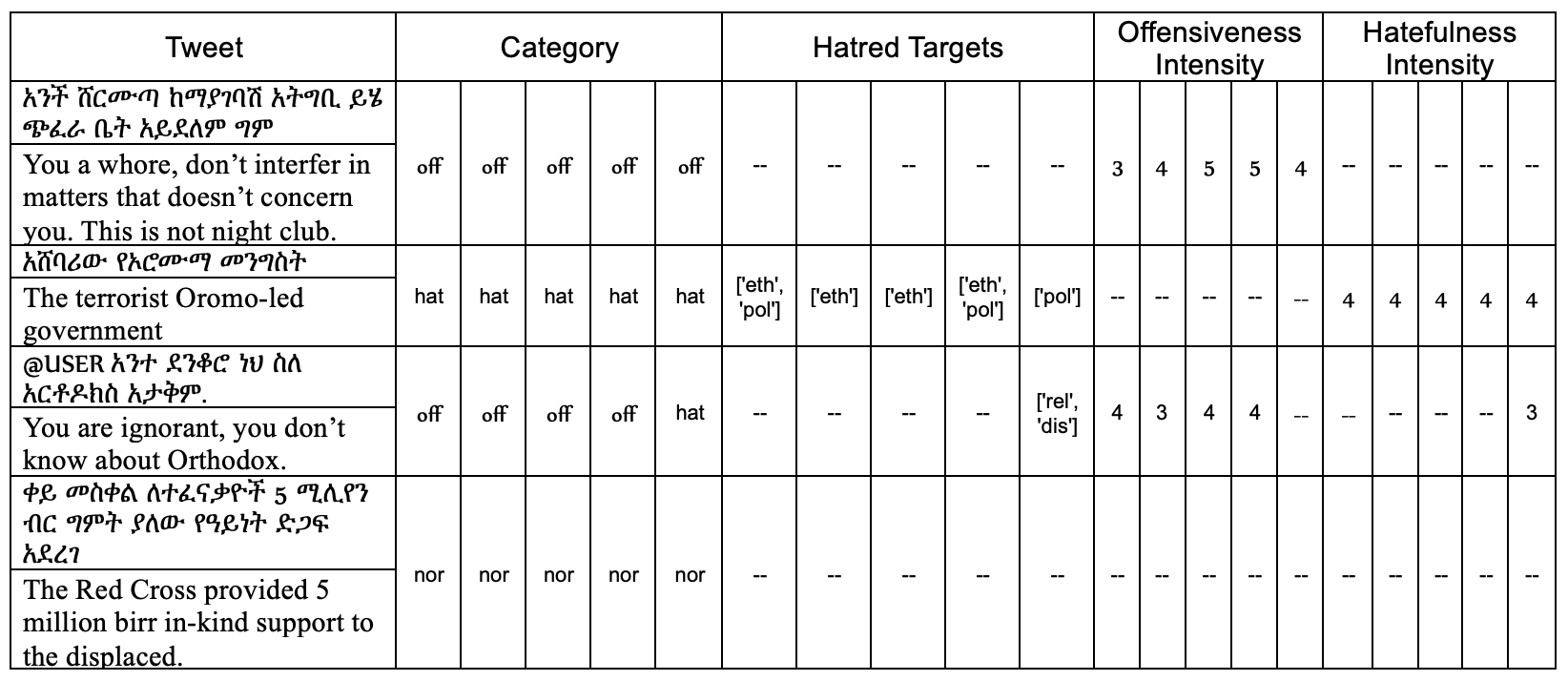} 
\caption{Dataset examples with 5 annotators for category, hatred target and  intensity (hatefulness and offensiveness) annotations. \textbf{Keys:} off = offensive, hat = hate, nor = normal, eth = ethnicity, pol = politics, rel = religion, dis = disability}
\label{tab-dataset_example}
\end{center}
\end{table*} 
\subsubsection{Tweet Category Annotation}

As indicated in Table \ref{label-distribution}, the 5 annotators absolutely agreed on 3.2k tweets out of 8.3k, which is 39\% of the total dataset. The absolute agreements on each category label among the annotators consisted of 38\% and 31\% for hateful and offensive tweets, respectively. The best absolute agreement of 49\% per category label is achieved for the normal class. The indeterminate class consisting of only 42 tweets, demonstrated exceptionally infrequent occurrence and is excluded from our experiments. The indeterminate tweets are composed in a language other than Amharic or are unintelligible, thus failing to convey clear messages to the annotators. While determining majority-voted tweets for two labels with equal frequency of 2, we handle ambiguities by giving priority to \textbf{hate, offensive, and indeterminate} labels, respectively. 

\begin{table}
\centering
\small
\begin{tabular}{lrrr}
\hline
\textbf{Label} & \makecell{\textbf{Majority} \\ \textbf{Voted}} & \makecell{\textbf{Fully}\\ \textbf{Agreed}} & \makecell{\textbf{Fully}\\ \textbf{Agreed \%}} \\
\hline
Hate & 4,149 & 1,575 & 38\% \\
Offensive & 2,164 & 664 & 31\% \\
Normal & 1,945 & 956 & 49\% \\
Indeterminate & 42 & 6 & 14\% \\
\hline
\textbf{Total} & \textbf{8,300} & \textbf{3,201} &  \textbf{39\%} \\
\hline
\end{tabular}
\caption{\label{label-distribution} Distribution of majority voted and fully agreed on category labels.}
\end{table}

\subsubsection{Target Annotation}

As indicated in Table \ref{target-distribution}, 
a significant majority of the target dataset, totaling 3,249 tweets (53.4\%), comprised of instances expressing hatred and hostility towards \textbf{political} targets. Political hatred tweets primarily centered on individuals based on their political ideologies, affiliations, or support for specific occasions. While ethnic hatred tweets presented the second majority, 38.8\% of hateful tweets, religious and other targets exhibited smaller proportions in the dataset. Annotators achieved better absolute agreements on \textbf{ethnic}, \textbf{political}, and \textbf{religious} hatred targets. Overall, there is complete consensus on 14.3\% of the hatred targets, which amounts to 867 instances within the target dataset. However, \textbf{gender} and other targets such as \textbf{disability} are scarcely represented in this dataset, which addresses \textbf{RQ-2}. The \textbf{none\_hate} represented tweets that do not contain any hateful content. 

\begin{table}
\centering
\small
\begin{tabular}{lrrr}
\hline
\textbf{Target} & \makecell{\textbf{Majority} \\ \textbf{Voted}} & \makecell{\textbf{Fully}\\ \textbf{Agreed}} & \makecell{\textbf{Fully}\\ \textbf{Agreed \%}} \\
\hline
Ethnic & \textbf{2,357} & \textbf{326} & 14\% \\
    Politics & \textbf{3,249} & \textbf{487} & 15\% \\
Religion & 359 & 54 & 15\% \\
Gender & 42 & 0 & 0\% \\
Other & 33 & 0 & 0\% \\
None\_Hate & 2,220 & 1,620 & 73\% \\
\hline
\textbf{Total} & \textbf{8,300} & \textbf{2,487} & \textbf{30\%} \\

\hline
\end{tabular}
\caption{\label{target-distribution} Distribution of hatred targets across majority voted and fully agreed tweets.}
\end{table}

Table \ref{targets-concided} demonstrated the number of times different distinct targets appeared simultaneously across the 5 annotators within the original dataset. It provided a detailed overview of the collective perspectives of these annotators regarding the simultaneous presence of distinct targets. The majority of overlapping occurrences that happened between \textbf{ethnic} and \textbf{political} targets in the dataset showed how \textit{ethnic and political hatred targets frequently intersect and overlap with one another}, emphasizing the complex relationship between these two targets.   
This overlap is likely a manifestation of Ethiopia's political landscape, which is primarily structured around ethnic divisions \cite{ghaderi2023impact}. In Ethiopia, most political parties are established based on ethnic affiliations. This underscores the intricate connection between ethnicity and political tensions in the nation's sociopolitical context, which addresses \textbf{RQ-3}. 

\begin{table}
\centering
\small
\begin{tabular}{lrr}
\hline
\textbf{Coexisted Targets} & \textbf{Frequency} & \textbf{Percent} \\
\hline
Ethnic, Politics & \textbf{3,290} & \textbf{83.0\%} \\
Religion, Ethnic & 291 & 7.3\% \\
Religion, Politics & 281 & 7.1\% \\
Ethnic, Politics, Religion & 101 & 2.6\% \\
\hline
Major Co-occurrences  & 3,963 & 100\% \\
\hline
\end{tabular}
\caption{\label{targets-concided} Main overlapping occurrences of targets.}
\end{table}

\subsubsection{Intensity Level Annotation}
\label{sec:likert-anno}
We have organized our intensity level annotation task into three distinct segments. \textbf{Normal} texts are assigned a score of \textbf{0}, waiving the need for intensity level annotations. The offensiveness scale spans from \textbf{less offensive (1)} to\textbf{ very offensive (5)}, utilizing a 5-point Likert scale for intensity level annotation. Similarly, the intensity of hatefulness is also rated on a 5-point Likert scale, ranging from \textbf{less hate (1)} to \textbf{very hate (5)}.

Table \ref{label-severity} presented the offensiveness and hatefulness intensities of tweets that appeared at least 2 times as offensive and hateful across the 5 annotators, respectively. 
Average offensiveness and hatefulness intensities on majority-voted tweets are lower than the absolutely agreed tweets. The majority voted tweets exhibit wider ranges of intensities for both offensiveness and hatefulness, 0.40-4.80 and 0.40-5.0, respectively. This indicated that hate and offensive annotated tweets in the dataset are represented in a spectrum of wider ranges. Therefore, hatefulness and offensiveness \textbf{are not simple binary measures}, rather they exist on \textbf{a continuum with varying degrees of intensity}.

In the category of completely agreed tweets, the range of offensiveness intensity spans from a minimum average intensity of 1.60 to a maximum average intensity of 4.80 per tweet. Meanwhile, in the case of hateful tweets, their hatefulness intensity encompasses intensities ranging from a minimum of 1.40 to a maximum of 5.0 across the subset of entirely agreed tweets. 
The wider intensity ranges and the cumulative average intensity values for offensiveness and hatefulness on the completely agreed tweets highlight the presence of varying degrees of intensity, even among tweets that have absolute agreements.

\begin{table}
\centering
\small
\begin{tabular}{lrr}
\hline
\multicolumn{3}{c}{\textbf{Majority Voted}} \\
\hline
\textbf{Label} & \textbf{Range} & \textbf{G-avg} \\
\hline
Hate & 0.4-5.0 & 2.48 \\
Offensive & 0.4-4.8 & 2.34 \\
\hline
\end{tabular}
\begin{tabular}{rr}
\hline
\multicolumn{2}{c}{\textbf{Fully Agreed}} \\
\hline
\textbf{Range} & \textbf{G-avg} \\
\hline
1.4-5.0 & 3.56 \\
1.6-4.8 & 3.66 \\
\hline
\end{tabular}
\caption{\label{label-severity} Hatefulness and offensiveness intensities. The "range" indicates the intensity ranges per tweet while "G-avg" shows the grand average intensities. \textbf{Keys:} G-avg = Grand Average.}
\end{table}




\subsection{Mapping Hate and Offensive Intensities}

\citet{Bahador2023} categorized hate speech into three major \textbf{stages}, namely 1) early warning, 2) dehumanization and demonization, and 3) violence and incitement. The \textbf{early warning} category starts with targeting \textbf{out-groups}\footnote{Out-groups are anyone who does not belong in the group but belongs to another group} to different types of negative speech that have less intensity. \textbf{Dehumanization and demonization} involve dehumanizing and demonizing the out-groups and their members, associating with subhuman or superhuman negative characters. The last category, \textbf{violence and incitement}  starts from the conceptual to the physical attacks and can result in more severe consequences such as incitement to violence and or even death against the out-groups under target.

Similarly, \citet{chandra-etal-2020-abuseanalyzer} classifies online abuse into three labels; 1) \textbf{biased attitude}, 2) \textbf{acts of bias and discrimination}, and 3) \textbf{violence and genocide}; to showcase the mild, moderate, and severe categories of abuse intensity.

The classification categories of \citet{Bahador2023} and \citet{chandra-etal-2020-abuseanalyzer} are employed to represent the hatefulness and offensiveness intensities of tweets as indicated in Table \ref{label-severity-stages}. 
We employed the revised rating scale described in Section \ref{sec:likert-anno} and represent offensiveness into three stage categories \cite{chandra-etal-2020-abuseanalyzer}, mild, moderate, and severe represented by 1-3, 4, and 5 rating scales, respectively. Similarly, the first category of hatefulness, early warning is represented from 1-3 ratings on the 5-point Likert scale. The second, dehumanizing and demonizing, and the third, incitement to violence categories are represented with scale 4 and scale 5, respectively. 

\begin{table}
\centering
\small
\begin{tabular}{llrrr}
\hline
\textbf{Label} & \makecell{\textbf{Average} \\ \textbf{Range}} & \makecell{\textbf{Stage}} & \makecell{\textbf{Tweet} \\ \textbf{Count}} & \textbf{ \%}\\
\hline
\multirow{4}{*}\textbf{Offensive} & [0.2 - 3.0) & Mild & \textbf{2,008} & \textbf{69\%} \\
                               & [3.0 - 4.0) & Moderate & 676 & 23\% \\
                               & [4.0 - 5.0] & Severe & 245 & 8\% \\
                               \hline
\multirow{4}{*}\textbf{Hate} & [0.2 - 3.0) & \makecell{{Early } \\ {Warning}} & \textbf{3,489} & \textbf{72\%} \\
                               & [3.0 - 4.0) & \makecell{{Dehuman-} \\ {ization}} & 808 & 17\% \\
                               & [4.0 - 5.0] & \makecell{{Voilence \&} \\ {Incitement}} & 528 & 11\% \\
\hline
\end{tabular}
\caption{\label{label-severity-stages} Hatefulness and offensiveness intensity ranges, and distribution of tweets across stages.}
\end{table}

As shown in Table \ref{label-severity-stages}, we carefully selected tweets labeled offensive at least by two annotators and the remainder labeled normal to explore the offensiveness intensity of tweets. Similarly, we did the same for hatefulness and analyzed the hatefulness and offensiveness intensities separately.
Offensive tweets that fall under the mild category, start from 0.2 minimum average intensity when only one of the annotators chooses offensive and rates its' offensiveness 1, and end at 3 maximum average intensity value. Tweets under this category comprised 69\% of the offensive tweets and are assumed to be less offending when compared with the other categories. Highly offending tweets constitute 8\% of the offensive tweets that present incitement or threats of violence against an individual while the moderate category accounts for 23\% of the tweets that dehumanize or demonize individuals. 

\begin{figure}[!ht]
\begin{center}
\includegraphics[scale=0.275]{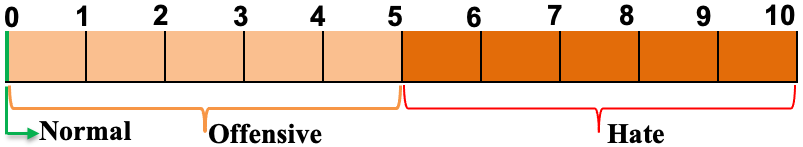}
\caption{Mapping the dataset in an 11-point Likert rating scale.}
\label{fig_data_mapping}
\end{center}
\end{figure}

The majority of hateful tweets comprised of 72\% tweets, fall under the less hate, early warning category. The 17\% and 11\% of tweets that fall under the second and third categories, respectively, require serious attention among different stakeholders such as the government, social media organizations, researchers, and non-governmental organizations (national and international). The mild and early warning stages of offensiveness and hatefulness can be taken as a demarcation point to enforce mitigation strategies by content moderators or other stakeholders. The playground for tackling hate and offensive speech on social media shall be at the first stages of early warning and mild, respectively.  
For our analysis and experimentation, we transform this scale to a range of 0 to 10, effectively creating an \textbf{11-point Likert scale}. In this revised scale, a score of 0 represents \textbf{normal} tweets while \textbf{offensive} and \textbf{hate} categories are scaled from 1 to 5 and 6-10 intensity ranges, respectively. The score of 1 and 5 denotes \textbf{less offensive} and \textbf{highly offensive} tweets, respectively. Similarly, 6 signifies \textbf{less hate}, and 10 represents a tweet characterized by \textbf{intense hate}. Figure \ref{fig_data_mapping} indicated the transformed dataset on an 11-point Likert rating scale. 
\subsection {Dataset Summary}
\begin{figure}[!ht]
\small
\begin{center}
\includegraphics[scale=0.4]{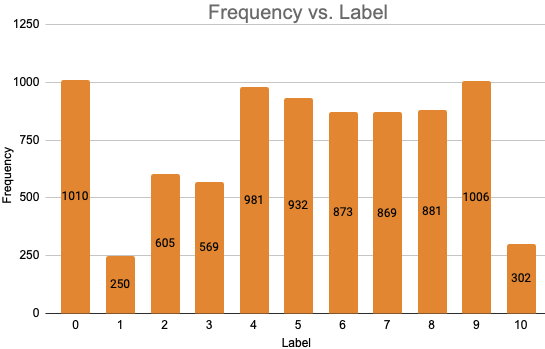}
\caption{Distributions of 0-10 rating labels.}
\label{fig_likert-label-distribution}
\end{center}
\end{figure}

A total of 8,258 instances were utilized for building classification and regression models, excluding the 42  indeterminate labeled instances. We presented the distributions of the dataset labels for the category, target, and intensity level classification and regression experiments in Table \ref{label-distribution}, Table \ref{target-distribution}, and Figure \ref{fig_likert-label-distribution}, respectively. 

We convert the average values calculated from the input of five annotators into whole numbers, resulting in a set of 11 labels spanning from 0 to 10. In this context, a label of 0 represents tweets labeled as \textbf{normal} while a label of 10 indicates tweets characterized as \textbf{extremely hateful}. Figure \ref{fig_likert-label-distribution} illustrates that scale labels 1 and 10 are associated with a relatively smaller number of instances in comparison to the other labels, as these values correspond to the two extremes of the spectrum.

\section{Experimental Setup}

We employed a 70:15:15 data-splitting approach to create the training, development, and test sets. This dataset remained consistent across all experiments, including \textbf{category classification}, \textbf{target classification}, and \textbf{intensity scale regression}. The development dataset was instrumental in refining the learning algorithms, and all the results reported in this study are based on data from the test set. 

We utilized the transformer models such as \textbf{AmRoBERTa}, \textbf{XLMR-Large-fintuned}, \textbf{AfroXLMR-large}, and \textbf{AfriBERTa} variants (small, base, large), and \textbf{AfroLM-Large (w/ AL) } for all experiments. AmRoBERTa is a RoBERTa-based language model that has been fine-tuned specifically with the Amharic language dataset, making it well-suited for downstream tasks and applications involving Amharic text \cite{fi13110275}. We also utilized Afro-XLMR-large \cite{alabi-etal-2022-adapting}, a multilingual language model tailored for African languages, including Amharic. This model demonstrated exceptional performance in various natural language processing tasks for African languages. Moreover, we fine-tuned the XLMR-Large \cite{xlmr2019} model using the same corpus that was utilized to train AmRoBERTa. We also employed the small, base, and large \textbf{AfriBERTa} variants \cite{ogueji-etal-2021-small}, and \textbf{AfroLM-Large (w/ AL)}, Pretrained multilingual models on many African languages including Amharic \cite{dossou-etal-2022-afrolm}. AfroLM Large (w/AL) is a special type of AfroLM Large which is designed with self active learning setups.

\section{Result and Discussion}


\begin{table}
\centering
\small
\begin{tabular}{lrrr}
\hline
\multicolumn{4}{c}{\textbf{Tweet category classification results (in \%)}} \\
\hline
\textbf{Classifier} &  \textbf{P}  & \textbf{R} & \textbf{F1}\\
\hline
AmRoBERTa & 75.01 & 75.06 &  74.82 \\
XLMR-large-finetuned & 73.60  & 73.45  & 73.50\\	
Afro-XLMR-large &  \textbf{75.37} & \textbf{75.30}  & \textbf{75.30}\\
AfriBERTa-large & 72.48 & 72.40 &  72.43 \\
AfriBERTa-base & 73.46 & 73.20 &  73.30 \\
AfriBERTa-small & 73.05 & 73.12 &  73.06 \\
AfroLM-Large (w/ AL)  & 72.02 & 71.99 &  71.98 \\
\hline
\multicolumn{4}{c}{\textbf{Hate target classification results (in \%)}} \\
\hline
AmRoBERTa &  66.74 & 66.42 & 66.02 \\
XLMR\_large\_fintuned &  65.57 &   66.18 &  65.85\\	
Afro\_XLMR\_large &  \textbf{70.34}   & \textbf{70.94}  & \textbf{70.59} \\
AfriBERTa\_large & 66.94 & 67.47 &  67.14 \\
AfriBERTa\_base & 66.04 & 66.42 &  66.11 \\
AfriBERTa\_small & 65.38 & 66.02 &  65.68 \\
AfroLM-Large (w/ AL)  & 64.26 & 64.57 &  64.23 \\
\hline
\end{tabular}
\caption{\label{tab:category-and-target-results}
Performance of models for category and hatred targets classification of tweets. \\
\textbf{Keys:} P = Precision, and R = Recall, AfroLM-Large (w/ AL) = AfroLM-Large (with Active Learning). 
}
\end{table}

As shown in Table \ref{tab:category-and-target-results}, the Afro-XLMR-large model outperformed the other 6 models on both tweet category and hatred target classification tasks with 75.30\% and 70.59\% F1-scores, respectively. In comparison to their performance on target classifications, all models exhibited a pronounced increase in all performance indicators such as precision, recall and F1-scores when undertaking the category classification task. Table \ref{tab:F1-score-varions} indicated the spectrum of F1-score variations across  diverse models. The performance variations observed in these two tasks extends from 4.71\% for Afro-XLMR-large to 8.80\% for AmRoBERTa. This disparity might be due to the class representation variations in the target classification task.
\begin{table}
\centering
\small
\begin{tabular}{lrrr}
\hline
\multicolumn{4}{c}{\textbf{F1-score variations across tasks (in \%)}} \\
\hline
\textbf{Classifier} &  \textbf{Cat.}  & \textbf{Tar.} & \textbf{Diff.}\\
\hline
AmRoBERTa & 74.82 & 66.02 & \textbf{8.80} \\
XLMR-large-finetuned  & 73.50 & 65.85 & 7.65\\	
Afro-XLMR-large &  \textbf{75.30} & \textbf{70.59} & \textbf{4.71} \\
AfriBERTa-large &  72.43 & 67.14 & 5.29 \\
AfriBERTa-base &  73.30 & 66.11 & 7.19 \\
AfriBERTa-small & 73.06 & 65.68 & 7.38 \\
AfroLM-Large (w/ AL)  & 71.98 & 64.23 & 7.75 \\
\hline
\end{tabular}
\caption{\label{tab:F1-score-varions}
F1-score Performance variations across models for category and hatred target classification tasks. \textbf{Keys:} Cat = Category, Tar = Target, and Diff = Difference, AfroLM-Large (w/ AL) = AfroLM-Large (with Active Learning).
}
\end{table}
\begin{table}[!ht]
\begin{center}
\small
\begin{tabular}{lr}
\hline
\multicolumn{2}{c}{\textbf{Regression results on Likert's 11-scale (in \%)}} \\
\hline
\textbf{Classifier} & \textbf{Pearson's cor. coeff. (r)} \\
\hline
AmRoBERTa & 77.23\\
XLMR-large-fintuned & 76.17\\	
Afro-XLMR-large & \textbf{80.22}\\
AfriBERTa\_large & 75.38 \\
AfriBERTa\_base & 76.57 \\
AfriBERTa\_small & 74.94 \\
AfroLM-Large (w/ AL)  & \textbf{80.22} \\
\hline
\end{tabular}
\caption{Performance of models on the regression tasks with Likert's 11-scale data.}
\label{tab:likert-classification-and-regression-results}
\end{center}
\end{table}



\begin{figure}[!ht]
\begin{center}
\includegraphics[scale=0.38]{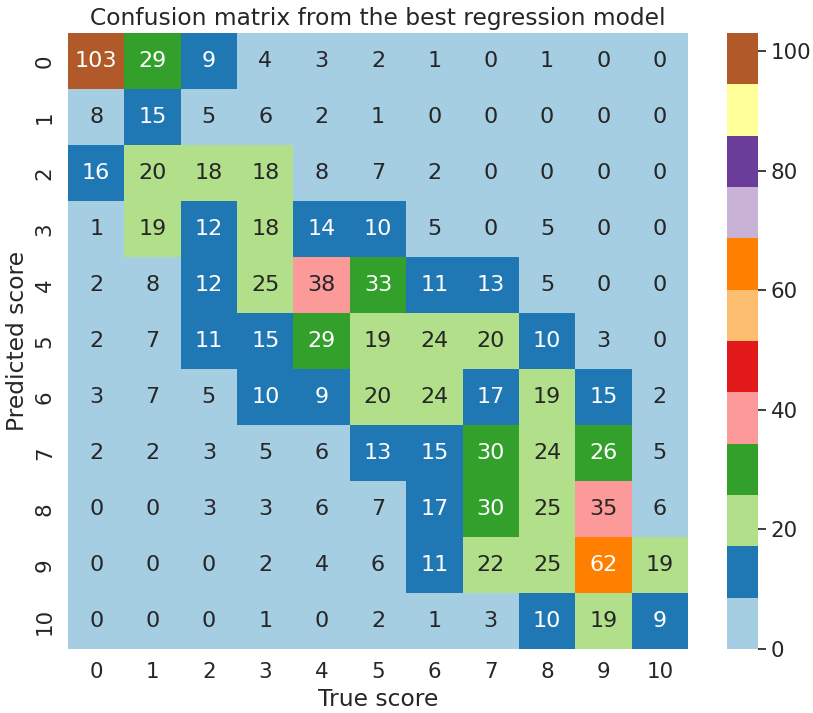}
\caption{Confusion matrix from Afro-XLMR-large.}
\label{fig-confusion-matrics-Regresion-model}
\end{center}
\end{figure}

We conducted \textbf{regression} experiments on the dataset collected through the utilization of an 11-point Likert scale, which was employed to measure intensity levels across a broad spectrum of ratings. In these experiments, real-valued scores spanning from 0 to 10 were utilized, and various models were applied for analysis. As part of our methodology, we focused on enhancing the visualization of the regression results for better interpretation. To achieve this goal, we rounded the results and illustrated them with visual representations presented in Figure \ref{fig-confusion-matrics-Regresion-model}.

Regression experiments were also performed on the 11-point Likert scale data with various models, and their performance was assessed using Pearson's r correlation coefficients. As suggested by \citet{schober2018correlation}, correlation coefficients falling between 0.70 and 0.89 are considered to indicate a strong correlation. Hence, the Pearson's r correlation coefficients achieved in this study, ranging from 74.94\% to 80.22\% demonstrated strong correlations. These findings denote a robust relationship between the predicted values and the actual observations, underscoring promising performance outcomes across all the models. The Afro-XLMR-large and AfroLM-Large (w/ AL)  models presented the best results in the intensity scaling regression tasks, which is 80.22\%.
%
%
%
%
%
Figure \ref{fig-confusion-matrics-Regresion-model} reveals that the majority of misclassified instances are clustered along the diagonal within the dark-colored boxes. This suggests that the true labels and their predicted counterparts are closely aligned. For instance, the true label 9 is frequently predicted as 7, 8, or 10, but seldom as 0, 1, 2, 3, or 4, which are considerably distant from 9. Conversely, there are only a few cases where extremely low true labels, such as 0, 1, 2, and 3, are predicted as higher extreme values, such as 7, 8, 9, or 10, and vice versa. In general, the regression model consistently displayed superior and more dependable performance as evidenced by the distribution of predictions in the confusion matrix.
The findings indicate that considering hate speech as a continuous variable, rather than adopting a binary classification, is a more suitable approach. Regression-based methods excel at capturing the intricate and evolving characteristics of hate speech, recognizing the subtle variations and intensities within this complex and sensitive domain.
This approach aligns with the dynamic and multifaceted nature of hate speech in the real-world situations, where it often exists on a spectrum of varying intensities, defying the usual simple binary categorization approaches. These findings address our research question, \textbf{RQ-1}.

\section{Conclusion and Future Work}

This paper introduced extensive benchmark datasets encompassing 8,258 tweets annotated for three tasks. These tasks included 1) \textbf{categorizing} hate speech into labels such as hate, offensive, and normal, 2) identifying the \textbf{targets} of hate speech, such as ethnicity, politics, and religion etc, and 3) assigning hate and offensive speech \textbf{intensity levels} using \textbf{Likert rating scales} to indicate offensiveness and hatefulness. To ensure robust annotation, each tweet is annotated by five annotators, resulting in a Fleiss kappa score of 0.49. Our contribution extended beyond the dataset itself; we provided comprehensive annotation guidelines tailored to each task and offered illustrative examples that effectively outlined the scope and application of these guidelines.
After a comprehensive analysis of the dataset, a clear pattern emerged, highlighting the prominence of \textbf{political} and \textbf{ethnic} targets, which mirrors the complex and unstable sociopolitical environment of Ethiopia. Notably, these two targets often co-occur in hateful tweets, underscoring the intricate nature of Ethiopia's sociopolitical dynamics, especially within ethnic contexts. Furthermore, our findings have demonstrated variations in the intensity of hate speech, emphasizing the necessity to develop regression models capable of gauging the level of toxicity in tweets.
We conducted a comprehensive exploration of various models for the detection of hate speech \textbf{categories}, their associated \textbf{targets}, and their \textbf{intensity levels}. Afro-XLMR-large demonstrated superior performance across all tasks \textbf{category classification}, \textbf{target classification} and \textbf{intensity prediction}. Our research illustrated that offensiveness and hatefulness cannot be simply categorized as binary concepts; instead, they manifest as continuous variables that assume diverse values along the continuum of ratings.

In the future, there is potential for a more in-depth examination of hatefulness and offensiveness intensities at finer levels. Moreover, the dataset could be subjected to further analysis to determine whether the predicted hate speech intensity levels can be employed as a valuable tool for monitoring and preventing potential conflicts, which would be particularly beneficial for peace-building efforts. We released our dataset, guidelines, top-performing models, and source code under a permissive license\footnote{\url{https://github.com/uhh-lt/AmharicHateSpeech}}.

\section*{Limitations}
The research study has the following limitations. The small dataset size, 8,258 tweets, could limit the robustness and applicability of the results to be generalized in various contexts. Secondly, the scarcity of the normal and offensive class instances within the dataset might impact the model's ability to accurately detect these categories. The extreme data imbalance in the target dataset, dominated by political and ethnic targets, might have affected the detection of other targets. 
The pre-selection strategy of tweets with dictionaries also affected the true distribution of hateful tweets in the corpus. Additionally, the smaller representations of label 1 and label 10 in the dataset annotated for rating intensity levels might have affected the performance of classification and regression models. These limitations collectively highlight the need for further investigations with larger datasets, and balanced representations of the examples for all the three types of tasks. 

\nocite{*}
\section{Bibliographical References}\label{sec:reference}

\bibliographystyle{lrec-coling2024-natbib}
\bibliography{lrec-colling2024}
\newpage
\section{Language Resource References}
\label{lr:ref}
\bibliographystylelanguageresource{lrec-coling2024-natbib}
\bibliographylanguageresource{languageresource}


\end{document}